\documentclass[conference]{IEEEtran}
\IEEEoverridecommandlockouts

\usepackage{cite}
\usepackage{amsmath,amssymb,amsfonts}
\usepackage{algorithmic}
\usepackage{graphicx}
\usepackage{textcomp}
\usepackage{xcolor}
\usepackage{makecell}
\usepackage{array}
\usepackage{booktabs}
\usepackage{multirow}
\usepackage{hyperref}

\def\BibTeX{{\rm B\kern-.05em{\sc i\kern-.025em b}\kern-.08em
    T\kern-.1667em\lower.7ex\hbox{E}\kern-.125emX}}

\begin{document}

\title{Safety-Critical Traffic Simulation with Guided Latent Diffusion Model}

\author{\IEEEauthorblockN{1\textsuperscript{st} Mingxing Peng}
\IEEEauthorblockA{\textit{The Hong Kong University of} \\
\textit{Science and Technology (Guangzhou)}\\
Guangzhou, China \\
mpeng060@connect.hkust-gz.edu.cn}
\and
\IEEEauthorblockN{2\textsuperscript{nd} Ruoyu Yao}
\IEEEauthorblockA{\textit{The Hong Kong University of} \\
\textit{Science and Technology (Guangzhou)}\\
Guangzhou, China \\
ryao092@connect.hkust-gz.edu.cn}
\and
\IEEEauthorblockN{3\textsuperscript{rd} Xusen Guo}
\IEEEauthorblockA{\textit{The Hong Kong University of} \\
\textit{Science and Technology (Guangzhou)}\\
Guangzhou, China \\
xguo796@connect.hkust-gz.edu.cn}
\and
\IEEEauthorblockN{4\textsuperscript{th} Yuting Xie}
\IEEEauthorblockA{\textit{School of Computer Science and Engineering} \\
\textit{Sun Yat-sen University}\\
Guangzhou, China \\
xieyt8@mail2.sysu.edu.cn}
\and
\IEEEauthorblockN{5\textsuperscript{th} Xianda Chen}
\IEEEauthorblockA{\textit{The Hong Kong University of} \\
\textit{Science and Technology (Guangzhou)}\\
Guangzhou, China \\
xchen595@connect.hkust-gz.edu.cn}
\and
\IEEEauthorblockN{\quad 6\textsuperscript{th} Jun Ma}
\IEEEauthorblockA{\quad \textit{The Hong Kong University of} \\
\textit{\quad Science and Technology (Guangzhou)}\\
\quad Guangzhou, China \\
\quad jun.ma@ust.hk}
\and
}

\maketitle

\begin{abstract}
Safety-critical traffic simulation plays a crucial role in evaluating autonomous driving systems under rare and challenging scenarios. However, existing approaches often generate unrealistic scenarios due to insufficient consideration of physical plausibility and suffer from low generation efficiency. To address these limitations, we propose a guided latent diffusion model (LDM) capable of generating physically realistic and adversarial safety-critical traffic scenarios. Specifically, our model employs a graph-based variational autoencoder (VAE) to learn a compact latent space that captures complex multi-agent interactions while improving computational efficiency. Within this latent space, the diffusion model performs the denoising process to produce realistic trajectories. To enable controllable and adversarial scenario generation, we introduce novel guidance objectives that drive the diffusion process toward producing adversarial and behaviorally realistic driving behaviors. Furthermore, we develop a sample selection module based on physical feasibility checks to further enhance the physical plausibility of the generated scenarios. Extensive experiments on the nuScenes dataset demonstrate that our method achieves superior adversarial effectiveness and generation efficiency compared to existing baselines while maintaining a high level of realism. Our work provides an effective tool for realistic safety-critical scenario simulation, paving the way for more robust evaluation of autonomous driving systems.
\end{abstract}


\section{Introduction}
\begin{figure}[!h]
\centerline{\includegraphics[width=\linewidth, trim= 0 0 0 0, clip]{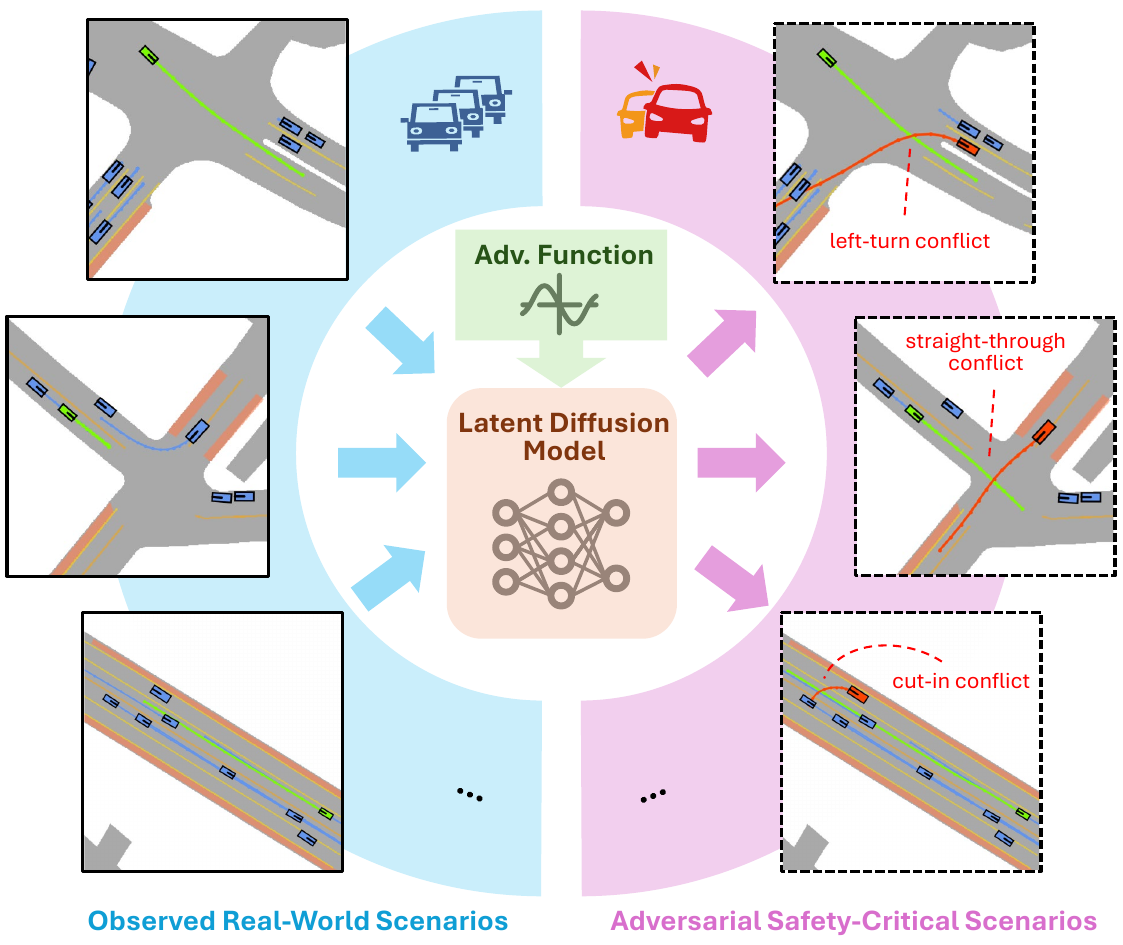}}
\caption{Overview of our guided LDM for safety-critical traffic simulation. Our model generates realistic yet adversarial driving scenarios by leveraging an LDM guided by adversarial objectives to effectively test AV systems.}
\label{fig:overview}
\end{figure}
With the rapid development of autonomous vehicle (AV) technology, rigorous and efficient evaluation of an AV’s safety performance has become a pivotal research focus, as reliable assessment is crucial for accelerating progress in autonomous driving systems \cite{guo2019safe, kang2019test}. However, in real-world road tests, traffic situations are predominantly nominal and safety-critical events occur only rarely, making the assembly of a sufficiently large corpus of such events both time-consuming and expensive. Consequently, safety-critical traffic simulation has become an indispensable component of the contemporary AV validation pipeline \cite{ding2023survey}.

Traditional approaches often rely on rule-based simulators such as CARLA \cite{dosovitskiy2017carla} and SUMO \cite{lopez2018microscopic}, which enable users to manually design safety-critical scenarios by explicitly specifying agent states. However, this manual scripting process demands substantial domain expertise and extensive parameter tuning. Even with careful manual design, the resulting scenes often fail to reproduce the multi-agent interactions and human-like driving behaviors present in natural traffic, thereby limiting their utility for large-scale evaluation. To overcome these limitations, recent data-driven methods first learn realistic driving behavior from large-scale trajectory datasets and then perform test-time optimization to synthesise safety-critical scenarios \cite{advsim, strive}. For example, Strive \cite{strive} employs a graph neural network (GNN) to construct a traffic prior for plausible traffic modeling and subsequently executes adversarial optimization at inference. Although such methods achieve moderate improvements in realism, they still struggle to model complex trajectory interactions accurately and lack guarantees of physical feasibility \cite{advsim, RN247}. Moreover, their iterative search procedures require substantial computational resources, resulting in low generation efficiency \cite{advsim, strive}.

Recently, diffusion models have shown remarkable performance in generating realistic sequential data, including interactive traffic scenarios that closely mirror real-world driving behaviors \cite{mao2023leapfrog, niedoba2024diffusion}. The iterative noise-to-data generation paradigm inherent in diffusion models captures complex multi-agent dependencies and enables flexible guidance during inference, thereby facilitating controllable generation \cite{zhong2023guided,jiang2023motiondiffuser}. Furthermore, recent advancements in latent diffusion models (LDMs) have demonstrated that learning compact and expressive latent representations of driving trajectories facilitates the modeling of their joint distributions, leading to improved realism and diversity in generated traffic scenarios \cite{xie2024advdiffuser, pronovost2023scenario}, while simultaneously enhancing computational efficiency \cite{rombach2022high}. Collectively, these works highlight the strong potential of diffusion models in generating realistic and controllable safety-critical scenarios.

Inspired by these works, we propose a guided LDM framework for simulating physically realistic and adversarial safety-critical traffic scenarios, as illustrated in Fig~\ref{fig:overview}. Specifically, we utilize a GNN-based encoder to transform both past and future trajectories into compact latent representations that capture intricate multi-agent interactions. Conditioned on the latent representation of past scene information, our model progressively denoises a noisy future latent to reconstruct realistic driving trajectories. 
To effectively guide the generation toward adversarial safety-critical scenarios, we introduce novel guidance objectives that progressively perturb the sampling process during inference. In addition, we design a sample selection module based on physical feasibility checks to ensure the generated scenarios adhere to physical constraints. The generated safety-critical scenarios are subsequently evaluated through closed-loop simulation using an ego vehicle equipped with a rule-based planner to assess their effectiveness and realism. This safety-critical traffic simulation framework facilitates exposing the limitations and vulnerabilities of autonomous driving systems, providing valuable insights for enhancing AVs' safety and robustness.

In summary, the main contributions of this paper include: 
\begin{itemize}
    \item We propose a guided LDM framework for safety-critical traffic simulation, which performs the denoising process in a compact latent space that effectively captures complex multi-agent interactions while improving computational efficiency.

    \item We design novel guidance objectives and a sample selection module based on physical feasibility checks, which jointly enable controllable generation of adversarial and physically realistic driving trajectories.

    \item Extensive experiments conducted on the nuScenes dataset demonstrate that our model significantly outperforms baseline methods in driving scenario generation in terms of adversarial effectiveness, diversity, and generation efficiency, while maintaining physically plausible and behaviorally realistic trajectories.
        
\end{itemize}

\section{Related Work}
\label{sec:related work}
This section reviews related works in two areas: safety-critical traffic simulation and diffusion-based traffic scenario generation.

\subsection{Safety-Critical Traffic Simulation} 
Safety-critical traffic simulation is crucial for evaluating autonomous driving systems under rare and high-risk scenarios~\cite{ding2023survey}. Traditional methods typically depend on simulators such as CARLA~\cite{dosovitskiy2017carla} and SUMO~\cite{lopez2018microscopic} to manually design such scenarios by modifying agent states. However, these approaches require significant domain expertise and often yield unrealistic or unscalable designs~\cite{waymo, metadrive}. To improve realism, recent research has leveraged large-scale driving datasets to learn realistic traffic patterns and generate safety-critical scenarios through optimization-based techniques at test time~\cite{advsim, strive, mixsim}. In particular, AdvSim~\cite{advsim} directly perturbs the standard trajectory space to induce adversarial trajectories, while Strive~\cite{strive} performs adversarial optimization in latent spaces using graph-based representations~\cite{scarselli2008graph}. MixSim~\cite{mixsim} learns route-conditioned policies to enable reactive and controllable re-simulation, supporting realistic variations and safety-critical interactions in mixed reality traffic scenarios. Although these approaches achieve moderate improvements in realism, they remain limited by inefficiency and continue to struggle with generating both plausible and controllable driving behaviors, thereby restricting their practicality for large-scale scenario generation.

\begin{figure*}[!h]
\centerline{\includegraphics[width=\linewidth, trim= 0 0 0 0, clip]{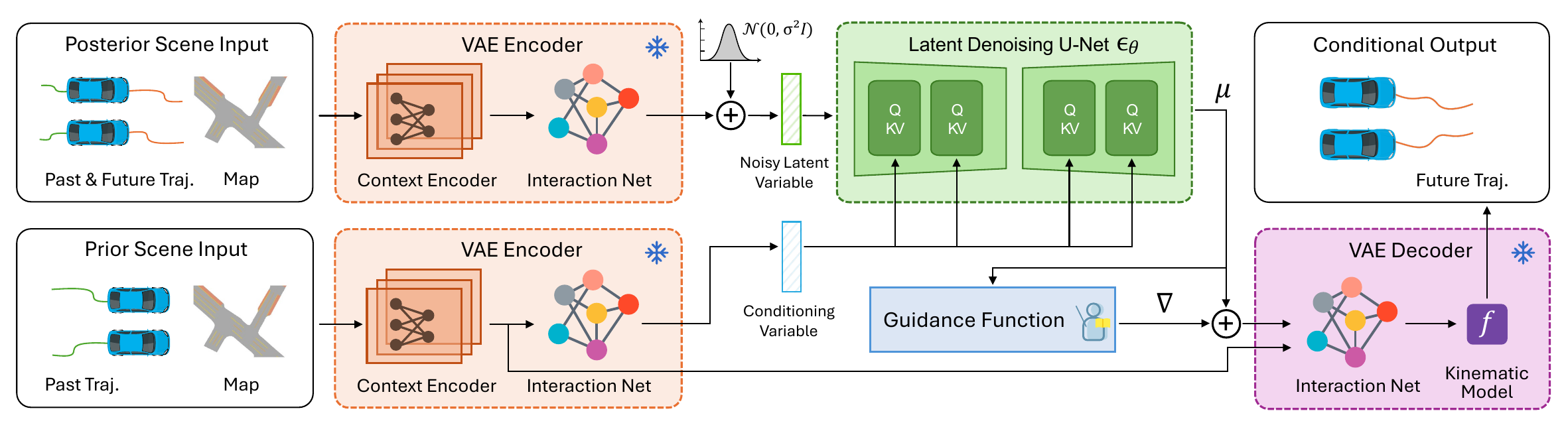}}
\caption{Overall framework of our proposed guided LDM. Graph-based VAEs encode posterior and prior scene input into latent representations, which are denoised by a U-Net conditioned on the prior latent. During the sampling stage, we introduce our proposed guidance objectives to guide the diffusion process toward generating adversarial safety-critical driving scenarios.}
\label{fig:overall_framework}
\end{figure*}

\subsection{Diffusion-based Traffic Scenario Generation}
Diffusion models \cite{ho2020denoising} have garnered significant attention due to their ability to model complex traffic patterns and generate high-fidelity simulations that closely resemble real-world scenarios. Moreover, they provide enhanced controllability, enabling the customization of traffic scenarios based on specific conditions or guidance \cite{peng2024diffusion}. For example, DJINN \cite{niedoba2024diffusion} utilized a classifier-free diffusion model to generate joint interactive trajectories for all agents in a traffic scene, conditioned on a flexible set of agent states. Additionally, several studies have further improved controllability by incorporating guidance mechanisms into the diffusion models during inference time \cite{zhong2023guided, jiang2023motiondiffuser, xu2023diffscene, chang2024safe, xie2024advdiffuser}. Specifically, CTG \cite{zhong2023guided} employs Signal Temporal Logic (STL) formulas as guidance for diffusion models to generate rule-compliant trajectories, while MotionDiffuser \cite{jiang2023motiondiffuser} proposes several differentiable cost functions as guidance, enabling the physical constraints in the generated trajectories. DiffScene \cite{xu2023diffscene} and Safe-Sim \cite{chang2024safe} proposed safety-based objective functions to simulate the safety-critical driving scenarios. Other studies have leveraged LDMs to learn more effective representations of driving trajectories and to model the joint distribution over these trajectories \cite{xie2024advdiffuser, pronovost2023scenario, chen2023executing}. However, these methods still face limitations in terms of controllability and flexibility. For example, AdvDiffuser \cite{xie2024advdiffuser} cannot be easily applied to generate scenarios for specific adversarial vehicles and requires the training of different classifiers for various adversarial strategies. 
\section{Methodology}
\label{sec:methodology}
This section presents our proposed model for generating realistic and adversarial safety-critical traffic scenarios. The overall framework of our proposed model is illustrated in Fig~\ref{fig:overall_framework}. We first provide a formal problem formulation of safety-critical scenario simulation. We then describe the LDM for scenario generation. Finally, we introduce novel guidance objectives and a physical feasibility-based sample selection module, which are used to guide the generation toward physically realistic and adversarial driving scenarios.

\subsection{Problem Formulation}
We focus on safety-critical traffic simulation involving $N$ agents, where one agent is the ego vehicle controlled by the planner $\pi$, and the future trajectories of the remaining $N-1$ vehicles are generated by our model. Among these vehicles, one is an adversarial vehicle whose goal is to cause a collision with the ego vehicle, while the others maintain realistic trajectories.

A driving scenario $S$ consists of $N$ agent states and a map $\textit{\textbf{m}}$. At each timestep $t$, the state of each agent $s^{i}_t = (x^{i}_t, y^{i}_t, \theta^{i}_t, v^{i}_t)$ represents the 2D position, heading, and speed. The corresponding actions of each agent are represented as $a^{i}_t = (\dot{v}i^t, \dot{\theta}i^t)$, indicating acceleration and yaw rate. We denote the past trajectories of all agents over the past $T_{hist}$ timesteps as $\textit{\textbf{x}} = \{s_{t-T_{hist}}, s_{t-T_{hist}+1}, ..., s_{t}\}$. The planner $\pi$ determines the ego vehicle's future trajectory over a time horizon from $t$ to $t + T$, denoted as $s^{0}_{t:t+T} = \pi (\textit{\textbf{m}}, \textit{\textbf{x}})$.

Our proposed LDM $g$, parameterized by $\theta$, simulates the future trajectories of the $N-1$ non-ego vehicles, denoted as $\boldsymbol{\tau} = \{s^{i}_{t:t+T}\}^{N-1}_{i=1}$. The model includes an encoder $\mathcal{E}$, which encodes historical trajectory data and map information into a compact latent representation, and a decoder $\mathcal{D}$, which decodes the denoised latent into the predicted future trajectories of the non-ego agents. During training, the model learns realistic traffic behaviors from real-world data, and during inference, adversarial objective functions guide the generation of safety-critical scenarios. 

\subsection{Latent Diffusion Models for Traffic Simulation}
We propose an LDM to generate realistic and controllable adversarial safety-critical driving scenarios through an iterative denoising process. In contrast to conventional diffusion methods that operate directly in trajectory space \cite{zhong2023guided, zhong2023language, chang2024safe}, our approach performs the denoising process in a latent space, thereby reducing computational overhead while enhancing feature expressiveness \cite{strive, xie2024advdiffuser, chen2023executing, rombach2022high}. Our model builds upon a pretrained graph-based VAE following Strive \cite{strive}, where the encoder captures complex multi-agent interactions and the decoder autoregressively generates future trajectories using a kinematic bicycle model to ensure plausibility of the generated trajectories.

\textbf{Architecture.}
As illustrated in Fig~\ref{fig:overall_framework}, our LDM consists of three components: two frozen GNN-based encoders, a learnable U-Net denoising network, and a frozen GNN-based decoder. The prior encoder $\mathcal{E}_\theta(\textit{\textbf{x}}, \textit{\textbf{m}})$ encodes past agent trajectories and local map features to produce the conditioning input $\textit{\textbf{c}}$, while the posterior encoder additionally incorporates future trajectories to yield the latent variable $\mathbf{z}$. The forward process begins from clean latent $\mathbf{z}^0 \sim q(\mathbf{z}^0)$, and progressively injects Gaussian noise through the following transition: 
\begin{equation} 
q(\mathbf{z}^k \mid \mathbf{z}^{k-1}) = \mathcal{N} \left(\mathbf{z}^k ; \sqrt{1 - \beta_k} \mathbf{z}^{k-1}, \beta_k \mathbf{I} \right) 
\end{equation} 
where $\beta_k$ denotes the predefined variance schedule controlling the noise level at each step. For a sufficiently large number of steps $K$, the distribution of $\mathbf{z}^K$ converges to an isotropic Gaussian, i.e., $\mathcal{N}(\mathbf{0}, \mathbf{I})$.

To accelerate inference, we adopt the Denoising Diffusion Implicit Models (DDIM) sampling strategy \cite{song2020denoising}, which enables non-Markovian reverse diffusion and supports efficient sampling by skipping intermediate steps without requiring retraining. The reverse process is defined as:
\begin{equation}
\mathbf{z}^{k-1} = \sqrt{\alpha_{k-1}} \cdot \tilde{\mathbf{z}}^0 + \sqrt{1 - \alpha_{k-1}} \cdot \epsilon_\theta(\mathbf{z}^k, k, \mathbf{c})
\label{denoising_process}
\end{equation}
where $\epsilon_\theta(\mathbf{z}^k, k, \textit{\textbf{c}})$ denotes the noise prediction model conditioned on $\textit{\textbf{c}}$, and $\alpha_k = \prod_{i=1}^{k} (1 - \beta_i)$ represents the cumulative product of the noise schedule. The predicted clean latent $\tilde{\mathbf{z}}^0$ is estimated as:
\begin{equation}
\tilde{\mathbf{z}}^0 = \left( \frac{\mathbf{z}^k - \sqrt{1 - \alpha_k} \cdot \epsilon_\theta(\mathbf{z}^k, k, \mathbf{c})}{\sqrt{\alpha_k}} \right)
\end{equation}
Iteratively applying this reverse process starting from $\mathbf{z}^K$ yields the final denoised latent $\hat{\mathbf{z}}^0$.

Finally, the decoder $\mathcal{D}_\theta(\hat{\mathbf{z}}^0, \textit{\textbf{x}}, \textit{\textbf{m}})$ autoregressively generates agent actions based on the denoised latent and the historical context, with the generated actions propagated through a kinematic bicycle model to ensure the physical plausibility of the resulting trajectories.

\textbf{Training.}
We fix the pretrained VAE and train only the LDM by minimizing the noise prediction loss: 
\begin{equation}
\mathcal{L} = \mathbb{E}_{\mathbf{z}^k, \epsilon \sim \mathcal{N}(0, I), k, \textit{\textbf{c}}} \left[ \|\epsilon - \epsilon_{\theta}(\mathbf{z}^k, k, \textit{\textbf{c}})\|^2 \right]
\end{equation}
where $\epsilon \sim \mathcal{N}(0, I)$ is the Gaussian noise, and $\epsilon_{\theta}(\mathbf{z}^k, k, \textit{\textbf{c}})$ is the noise prediction model, which is conditioned on the past context latent $\textit{\textbf{c}}$. 

\subsection{Controllable Generation with Guidance Function}
To enable the controllable generation of safety-critical driving scenarios, we introduce an objective function $\mathcal{J}(\boldsymbol{\tau})$, which guides the denoising process. Since the denoising process operates in the latent space, each guided iteration step first decodes the latent vector $\mathbf{z}$ into the corresponding trajectory $\boldsymbol{\tau}$ using a decoder $\mathcal{D}$, formulated as $\boldsymbol{\tau} = \mathcal{D}_\theta (\mathbf{z}, \textit{\textbf{x}}, \textit{\textbf{m}})$. At each reverse step $t$, the gradient of the guidance objective is injected into the predicted noise following \cite{dhariwal2021diffusion}:
\begin{equation}
\tilde{\boldsymbol{\epsilon}}_\theta
=
\boldsymbol{\epsilon}_\theta
-
s\,\nabla_{\mathbf{z}_t}\mathcal{J}(\mathcal{D}_\theta (\mathbf{z}^t, \textit{\textbf{x}}, \textit{\textbf{m}})),
\label{eq:guided_noise}
\end{equation}
where $s$ denotes the guidance scale. The perturbed latent $\mathbf{z}^{k-1}$ is then updated using the DDIM formulation defined in (\ref{denoising_process})

In detail, the objective $\mathcal{J}(\boldsymbol{\tau})$ is a combination of three components: 
\begin{equation}
\mathcal{J}(\boldsymbol{\tau})= 
w_{b}\mathcal{J}_{\mathrm{br}}(\boldsymbol{\tau})\;
+\;
w_{ar}\mathcal{J}_{\mathrm{ar}}(\boldsymbol{\tau})\;
+\;
w_{a}\mathcal{J}_{\mathrm{adv}}(\boldsymbol{\tau})
\end{equation}
where $\mathcal{J}_{br}$ is a realism constraint for non-adversarial vehicles, ensuring them to avoid collisions with each other and prevent off-road deviations. Similarly, $\mathcal{J}_{ar}$ is a realism constraint for adversarial vehicles, ensuring that adversarial agents maintain plausible behaviors without colliding with non-adversarial vehicles or leaving the roadway. The adversarial objective $\mathcal{J}_{adv}$ controls the adversarial vehicle’s behavior to induce a collision with the ego vehicle. The $w_b$, $w_ar$, and $w_a$ are three hyperparameters controlling the weights of three different objectives.

The collision penalty between vehicles is defined as follows:
\begin{equation}
\text{veh\_coll\_pens}_{ij}(t) =
\begin{cases} 
1 - \frac{d_{ij}(t)}{p_{ij}}, & \text{if } d_{ij}(t) \leq p_{ij} \\
0, & \text{otherwise}
\end{cases}
\end{equation}
where $d_{ij}(t)$ denotes the Euclidean distance between vehicles $i$ and $j$ at time $t$, and $p_{ij} = r_i + r_j + d_{\text{buffer}}$ represents the collision threshold determined by the sum of the vehicle radii and a predefined safety buffer.

Similarly, the map collision penalty is formulated as:
\begin{equation}
\text{env\_coll\_pens}_{i}(t) =
\begin{cases} 
1 - \frac{d_{i}(t)}{p_{i}}, & \text{if } d_{i}(t) \leq p_{i} \\
0, & \text{otherwise}
\end{cases}
\end{equation}
where $d_{i}(t)$ denotes the distance from the vehicle center to the nearest non-drivable area at time $t$, and $p_{i}$ denotes the maximum allowable displacement before a collision.

In practice, $\mathcal{J}_{br}$ and $\mathcal{J}_{ar}$ are computed by accumulating the respective vehicle collision and map collision penalties over all relevant agents and timesteps. In contrast, the adversarial objective is defined as: 
\begin{equation}
\mathcal{J}_{\mathrm{adv}}(\boldsymbol{\tau}) = \sum_{t=1}^{T} min(0,\ d(t)\;-\;p)
\end{equation}
where \(d(t)\) is the current center-to-center distance between the adversarial vehicle and ego vehicle at time \(t\), and \(p\) denotes the collision threshold between the adversarial vehicle and the ego vehicle. Accordingly, these three guidance objectives are computed separately for non-adversarial and adversarial vehicles, with corresponding latent variables updated independently. Specifically, the gradient of $\mathcal{J}_{br}$ is applied to the latent representations of non-adversarial vehicles, while the gradient of the combined objective comprising $\mathcal{J}_{ar}$ and $\mathcal{J}_{adv}$ is used to update the latent variables corresponding to the adversarial vehicle, thereby ensuring targeted control over its behavior.

\textbf{Sample Selection Module.}  
At inference time, we generate multiple candidate future trajectories for the non-ego agents in a given scene and rank them using a weighted score:
\begin{equation}
\mathcal{C} =
w_g\,\mathcal{J}(\boldsymbol{\tau})
+ w_p\,(1 - \Phi(\boldsymbol{\tau})),
\end{equation}
where \( \Phi(\boldsymbol{\tau}) \) is a physical feasibility indicator function defined as:
\begin{equation}
\Phi(\boldsymbol{\tau}) = a_{\text{lon}}(\boldsymbol{\tau}) \wedge a_{\text{lat}}(\boldsymbol{\tau}),
\end{equation}
with \( a_{\text{lon}}(\boldsymbol{\tau}) \) indicating compliance with longitudinal acceleration constraints and \( a_{\text{lat}}(\boldsymbol{\tau}) \) indicating compliance with lateral acceleration constraints. The candidate trajectory with the lowest score \( \mathcal{C} \) is selected.

\section{Experimental Results}
\label{sec:experiment}
This section presents experimental results validating the effectiveness of our proposed model. We first introduce the dataset and evaluation metrics used in our study. We then compare our method with representative baselines on both real traffic simulation and safety-critical scenario generation tasks. Furthermore, we conduct an ablation study to analyze the contribution of the proposed guidance mechanism.

\begin{table*}
  \caption{Comparison of real traffic simulation results on the nuScenes dataset. We compare our diffusion model without guidance against AdvSim \cite{advsim} and Strive \cite{strive}, both evaluated without adversarial optimization.}
  \centering
  \renewcommand{\arraystretch}{1.5} 
  \setlength{\tabcolsep}{13pt}
  \begin{tabular}{c|ccccc|c}
    \toprule 
    \centering \multirow{2}{*}{Model} & \multicolumn{5}{c|}{Realism} & Diversity \\
    & Veh Coll (\%) $\downarrow$ & Env Coll $\downarrow$ & ADE (m) $\downarrow$  & FDE (m) $\downarrow$ & minSFDE (m) $\downarrow$ 
    & FDD (m) $\uparrow$ \\
    \midrule
    AdvSim  & 0.78 & \textbf{14.92} & \textbf{2.24} & \textbf{5.39} & 5.39 & 0.0 \\
    Strive  & 0.36 & 16.23 & 2.74 & 6.65 & 3.68 & \textbf{13.82} \\
    Ours  & \textbf{0.31} & 15.99 & 2.41 & 5.94 & \textbf{3.66} & 10.25 \\
    \bottomrule
  \end{tabular}
  \label{tab:real_traffic}
\end{table*}

\begin{table*}
  \caption{Comparison of safety-critical traffic simulation results on the nuScenes dataset. We compare our approach against AdvSim \cite{advsim} and Strive \cite{strive} for safety-critical traffic simulation with a rule-based planner.}
  \centering
  \renewcommand{\arraystretch}{1.5} 
  \setlength{\tabcolsep}{3.5pt}
  \begin{tabular}{c|cc|cccccc|c}
    \toprule 
    \centering \multirow{2}{*}[-1.0ex]{Model} & \multicolumn{2}{c|}{Adversariality} & \multicolumn{6}{c|}{Realism} & Efficiency \\
    & \makecell{Adv-Ego \\ Coll (\%) $\uparrow$} & Adv Acc $\uparrow$  
    & \makecell{Adv \\ Offroad (\%) $\downarrow$}  
    & \makecell{Other \\ Offroad (\%) $\downarrow$}  
    & \makecell{Adv-Other \\ Coll (\%) $\downarrow$}   
    & \makecell{Other-Ego \\ Coll (\%) $\downarrow$}  
    & \makecell{Other-Other \\ Coll (\%) $\downarrow$} 
    & Other Acc $\downarrow$ 
    & Infer time (s) $\downarrow$ \\
    \midrule
    AdvSim  & 24.72 & 0.90 & 15.60 & \textbf{14.85} & \textbf{0.56} & \textbf{0.91} & 0.11 & 0.38 & 338.35 \\
    Strive  & 22.69 & 0.88 & 18.94 & 16.64 & 0.90 & 1.08 & \textbf{0.05} & 0.39 & 609.72\\
    Ours    & \textbf{38.17} & \textbf{1.14} & \textbf{11.49} & 16.83 & 5.68 & 1.47 & 0.63 & \textbf{0.33} & \textbf{171.60} \\
    \bottomrule
  \end{tabular}
  \label{tab:sc_traffic}
\end{table*}

\subsection{Dataset}
We conduct our experiments on the nuScenes dataset~\cite{nuscenes2019}, which consists of 1,000 driving scenes, each lasting 20 seconds at 2,Hz. Models are trained on the provided training split and evaluated on the validation split. Following the nuScenes prediction challenge guidelines, we use 2 seconds (4 timesteps) of past motion to predict the next 6 seconds (12 timesteps).

\subsection{Evaluation Metrics}
We evaluate the generated scenarios from three perspectives: adversariality, realism, and diversity. Adversariality measures the effectiveness in creating safety-critical scenarios for the ego vehicle using metrics such as Adv-Ego Collision Rate and Adv Acceleration, with higher values indicating stronger adversarial behavior. Realism assesses the plausibility of the scenarios through metrics like vehicle collision rates (Veh Coll) and map collision rates (Env Coll), further subdivided into collision rates among different agent types (Adv-Other, Other-Ego, Other-Other) and map collision rates among different agent types (Adv Offroad, Other Offroad), where lower values reflect more realistic behavior. Diversity is quantified using the Final Displacement Diversity (FDD)~\cite{mixsim, xie2024advdiffuser}, with higher FDD values indicating greater scenario variation. Additionally, we employ metrics such as Average Displacement Error (ADE), Final Displacement Error (FDE), and minimum Scenario Final Displacement Error (minSFDE) to measure overall trajectory accuracy, along with inference time to evaluate generation efficiency.

\subsection{Experimental Setup}
\textbf{Baselines.} We compare our method with two representative baselines. AdvSim\cite{advsim} optimizes the acceleration of a predefined adversarial vehicle to induce collisions, with initial states generated by SimNet\cite{bergamini2021simnet}. Strive~\cite{strive}, using its official implementation, conducts adversarial optimization in a learned latent space based on a traffic model.

\textbf{Implementation Details.} Our model is implemented in PyTorch and trained on four NVIDIA RTX 4090 GPUs for six hours. The diffusion model is trained for 200 epochs using Adam with a learning rate of $5 \times 10^{-4}$, 20 diffusion steps, and 10 test samples.

To ensure fair comparison in controllability, the adversarial vehicle is selected as the one closest to the ego vehicle in the initial state and must satisfy feasibility constraints \cite{strive}. Unlike Strive, where the adversarial vehicle may change dynamically, we fix the selected adversarial agent throughout the scenario. Moreover, in all experiments, the ego vehicle is controlled by a rule-based planner to ensure consistency across different methods.

\subsection{Simulating Real Traffic}
To evaluate the effectiveness of our LDM in simulating real-world traffic scenarios, we conduct comparative experiments on the nuScenes dataset. Specifically, we compare our unguided diffusion-based model against two representative baselines: the VAE-based Strive model and the imitation learning-based AdvSim model, both evaluated without adversarial optimization. As shown in Table~\ref{tab:real_traffic}, our model achieves both realistic and diverse trajectory generation, demonstrating its ability to capture complex traffic behaviors while maintaining diversity. In particular, our model exhibits stronger plausibility, as evidenced by a lower vehicle collision rate (0.31\%) compared to both AdvSim (0.78\%) and Strive (0.36\%). This result indicates that our approach more effectively captures multi-agent interactions and generates coherent traffic behaviors. Furthermore, unlike the deterministic generation process of AdvSim, both Strive and our model employ sampling-based generative processes, inherently supporting more diverse trajectory generation. Notably, our model achieves competitive diversity while simultaneously preserving realism, demonstrating its ability to simultaneously enhance both fidelity and flexibility in scenario synthesis.

\subsection{Simulating Safety-Critical Scenarios}
Comparison of safety-critical traffic simulation results is shown in Table~\ref{tab:sc_traffic}. Compared to the baselines, our proposed model demonstrates significant advantages in generating adversarial scenarios while maintaining a high level of realism and ensuring efficient generation. Our model achieves the highest adversarial effectiveness, with an \textit{Adv-Ego Collision Rate} of 38.17\%, substantially surpassing AdvSim (24.72\%) and Strive (22.69\%). This indicates the model's strong ability to generate safety-critical scenarios that effectively challenge autonomous driving systems. Meanwhile, the adversarial behaviors remain behaviorally plausible. The off-road rate of the adversarial vehicle is only 11.49\%, significantly lower than that of AdvSim (15.60\%) and Strive (18.94\%), suggesting that our model generates realistic yet aggressive behaviors without violating road constraints. In terms of generation efficiency, our diffusion-based approach achieves an average inference time of 171.60 seconds, which is markedly faster than test-time optimization-based methods such as AdvSim (338.35 s) and Strive (609.72 s). 
Overall, our approach achieves strong adversarial performance and concurrently maintains high level realism and generation efficiency, making it well-suited for large-scale safety validation of AVs.

\subsection{Ablation Study}
\begin{figure}[t]
    \centering
    \includegraphics [width=\linewidth, trim=5 10 0 0, clip]{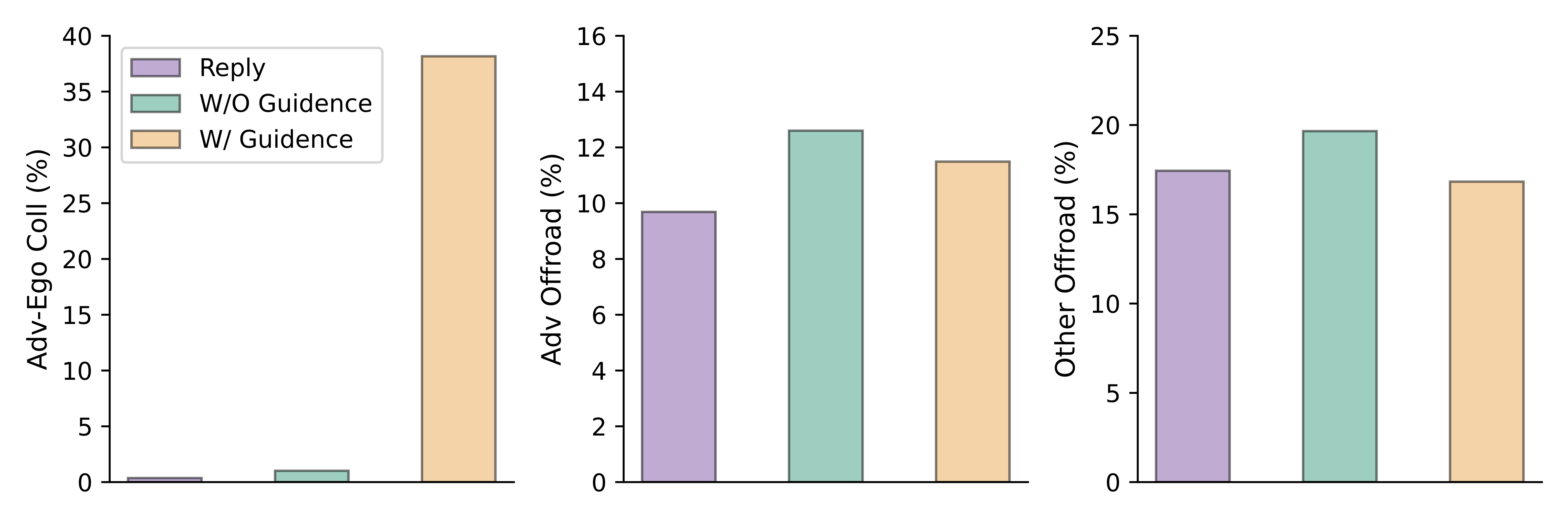}
    \caption{Ablation study on guidance components. \textit{Reply} refers to simulation using replayed non-ego vehicle trajectories from the dataset. \textit{W/O Guidance} denotes our diffusion-based model without guidance, and \textit{W/ Guidance} represents our proposed guidance-based diffusion model.}
    \label{fig:ablation_study}
\end{figure}

To investigate the effectiveness of the proposed guidance mechanism, we perform an ablation study comparing three settings: \textit{Reply}, which replays the original non-ego vehicle trajectories from the dataset; \textit{W/O Guidance}, which runs our diffusion-based model without guidance; and \textit{W/ Guidance}, which includes the proposed guidance module in the diffusion process.

As shown in Fig.~\ref{fig:ablation_study}, the \textit{Reply} setting yields minimal adversarial effectiveness, with a near-zero \textit{Adv-Ego Collision Rate}, as agents simply replay dataset trajectories. The results of \textit{W/O Guidance} model demonstrate a strong ability to simulate realistic traffic scenarios, achieving plausible vehicle behaviors and moderate off-road rates. With the integration of the guidance module, the \textit{W/ Guidance} model further improves adversariality, achieving the highest Adv-Ego Collision Rate, demonstrating the effectiveness of adversarial guidance. Moreover, Adv Offroad and Other Offroad are slightly reduced, indicating that the real-based guidance contributes to improving the realism of generated scenarios without degrading adversarial effectiveness. These findings demonstrate that the proposed guidance mechanism plays a critical role in generating safety-critical scenarios while maintaining a high level of realism.
\section{Conclusion}
\label{sec:conclusion}
In this paper, we introduce a guided LDM framework for simulating safety-critical traffic scenarios. By performing diffusion in a compact latent space learned from a graph-based VAE, our approach effectively captures complex multi-agent interactions while improving computational efficiency. To enable controllable generation of safety-critical scenarios, we propose innovative guidance objectives that guide the diffusion process to generate adversarial and behaviorally realistic driving behaviors. Additionally, we introduce a simple yet effective sample selection module based on physical feasibility checks, which further enhances the physical plausibility of the generated scenarios. Extensive experiments on the nuScenes dataset demonstrate that our method outperforms existing baselines in terms of adversarial effectiveness, realism, diversity, and generation efficiency. In the future, we plan to explore integrating large language models (LLMs) or AI agents to further enhance the controllability and performance of safety-critical traffic simulation.

\bibliographystyle{ieeetr}
\bibliography{reference}

\end{document}